\documentclass[a4paper,11pt]{article}
\usepackage{jinstpub_1} % for details on the use of the package, please
\title{\boldmath Preference$-$based performance measures for Time$-$Domain Global Similarity method}

\author[a]{Ting. Lan,}
\author[a, 1]{Jian. Liu,\note{Corresponding author.}}
\author[a, b]{Hong. Qin,}
\affiliation[a]{School of Nuclear Science and Technology and Department of Modern Physics, University of Science and Technology of China, Hefei, Anhui 230026, China}
\affiliation[b]{Plasma Physics Laboratory, Princeton University, Princeton, NJ 08543, USA}
\emailAdd{jliuphy@ustc.edu.cn}

\abstract{For Time$-$Domain Global Similarity (TDGS) method, which transforms the data cleaning problem into a binary classification problem about the physical similarity between channels, directly adopting common performance measures could only guarantee the performance for physical similarity. Nevertheless, practical data cleaning tasks have preferences for the correctness of original data sequences. To obtain the general expressions of performance measures based on the preferences of tasks, the mapping relations between performance of TDGS method about physical similarity and correctness of data sequences are investigated by probability theory in this paper. Performance measures for TDGS method in several common data cleaning tasks are set. Cases when these preference$-$based performance measures could be simplified are introduced. }

\begin{document}
\maketitle
\flushbottom

\section{Introduction}
\label{sec:intro}

To guarantee the availability and reliability of data source, a general-purposed Time$-$Domain Global Similarity (TDGS) method based on machine learning techniques has been developed, which sorts out the incorrect fusion data by classifying the physical similarity between channels~\cite{2017arXiv170504947L}. In the model selection and evaluation process of TDGS method, different performance measures lead to models of various generalization abilities~\cite{karayiannis2013artificial,kohavi1995study}. Choices of performance measures depend on the required generalization ability of models, or say preferences of tasks. Setting preference$-$based performance measures helps to perform corresponding tasks better. For TDGS method, directly adopting common performance measures, such as precision, recall, F$-$factor, confusion matrix, and Receiver Operating Characteristics (ROC) graphs, could only guarantee the performance for physical similarity between data sequences~\cite{goutte2005probabilistic,powers2011evaluation,fawcett2006introduction}. Nevertheless, practical data cleaning tasks have requirements for the correctness of original data sequences. For example, some data cleaning tasks require high recall rate of incorrect data, and some tasks require high precision of correct data. To improve the performance of TDGS method in data cleaning tasks, new performance measures based on the preferences of corresponding tasks should be set.

Each sample of TDGS method is the combination of two data sequences from different channels of MUlti-channel Measurement (MUM) systems. By tagging the sample completely constituted by correct data as physical similarity, and tagging the sample containing at least one incorrect data sequence as physical dissimilarity, the data cleaning problem turns into a binary classification problem about physical similarity between data sequences. When defining the prediction performance of TDGS method, True Positive (TP) refers that predicting results and actual sample tags are both dissimilar. True Negative (TN) refers that predicting results and actual sample tags are both similar. However, when defining the required prediction performance for data cleaning tasks, TP and TN refer to the incorrect and correct sequences which are correctly predicted. To set performance measures according to the preferences of tasks, the mapping relations between performance of TDGS method about physical similarity and correctness of data sequences should be explicit first. However, these mapping relations are complex and influenced by many factors, such as the data structure of samples, performance of models, the rule for judging the correctness of data based on given physical similarity, and the judging order. To obtain the general expression of preference$-$based performance measures for TDGS, the mapping relations between performance of TDGS method about physical similarity and correctness of data sequences are investigated by probability theory in this paper. Based on these mapping relations, we set preference$-$based performance measures for several common data cleaning tasks. By adopting these new performance measures in the model selection and evaluation process, models generated by TDGS method could best meet the preferences of tasks in probability.

The mapping relations between performance of TDGS method about physical similarity and correctness of data sequences are decided by the rules for judging the correctness of data based on given physical similarity. Here we adopt an absolute algorithm, i.e., by scanning through all samples tagged with similarity first, tag the sequences contained in the similar samples as correct data, and tag the other data as incorrect data. Based on this judging rule, the mapping relations between performance about physical similarity and correctness of data sequences can be analyzed by probability theory. In view that every prediction about physical similarity is independent of each other, the probability of judging the correctness of data is the product of the probabilities of all predictions employed in the judging process~\cite{durrett2010probability}. For example, according to the adopted judging rule, a correct data sequence $\ensuremath{S_0}$ would be predicted as incorrect if all samples containing $\ensuremath{S_0}$ are predicted as dissimilarity. Therefore, the probability of judging a correct data sequence as incorrect can be decided according to the number of similar samples containing $\ensuremath{S_0}$, the probability of predicting similar samples as dissimilarity, the number of dissimilar samples containing $\ensuremath{S_0}$, and the probability of predicting dissimilar samples as dissimilarity. Based on the mapping relations between performance of TDGS method about physical similarity and the correctness of data, performance measures for several common data cleaning tasks are set in this paper. Meanwhile, the correlative relations between these preference$-$based performance measures and performance parameters about physical similarity are analyzed. When preference-based performance measures are strong positive correlative with certain parameter, these performance measures could be simplified.

The rest parts of this paper are organized as follows. In section~\ref{sec:2}, the mapping relations between performance of TDGS method about physical similarity and the correctness of data sequences are studied by probability theory. In section~\ref{sec:3}, performance measures for several common data cleaning tasks are investigated. Cases when these performance measures could be simplified are introduced. In section~\ref{sec:4}, further optimizations of setting preference-based performance measures for TDGS method are discussed.

\section{Mapping relations between performance of TDGS method about physical similarity and correctness of data sequences}
\label{sec:2}

In this section, the correctness of data sequences based on performance of TDGS method about physical similarity is analyzed by probability theory. Corresponding mapping relations are explicitly exhibited. 

MUM system measures related yet distinct aspects of the same observed object with multiple independent measuring channels. Interferometer systems~\cite{kawahata1999far}, polarimeter systems~\cite{donne2004poloidal,brower2001multichannel,liu2014faraday,liu2016internal,zou2016optical}, and electron cyclotron emission imaging systems~\cite{luo2014quasi} are all typical MUM systems in Magnetic Confinement Fusion (MCF) devices. For practical purpose of data cleaning in MCF devices, the samples of a validation set are generated from diagnostic data of one discharge. For an N$-$channel MUM system, suppose $\ensuremath{n}$  and   are ${Q_1} = \left. {n} \middle/ {N} \right.$ the number and proportion of correct data sequences respectively. By combining two data sequences from different channels of MUM system as one sample,  $C_N^2$ samples can be generated. Among them, $C_n^2$ samples are similar, and  $C_N^2 - C_n^2$ samples are dissimilar. The prediction performance of TDGS method about physical similarity can be divided as four types. ${k_1}$  and ${k_2}$  are the probabilities of correctly and incorrectly predicting similar samples respectively. ${k_3}$  and ${k_4}$  are the probabilities of correctly and incorrectly predicting dissimilar samples respectively. The total probability of all predictions equals 1, i.e., $\sum\limits_{i = 1}^4 {{k_i}}  = 1$. The recall rate of similar samples ${Q_2}$ and the recall rate of dissimilar samples ${Q_3}$ are typical performance parameters about physical similarity, which are defined as the fraction of correctly predicted samples over total samples, namely
\begin{subequations}\label{eq:1}
\begin{align}
\label{eq:1:1}
\ensuremath{{Q_2} &= \frac{{{k_1}}}{{{k_1} + {k_2}}},}
\\
\label{eq:1:2}
\ensuremath{{Q_3} &= \frac{{{k_3}}}{{{k_3} + {k_4}}}.}
\end{align}
\end{subequations}

The proportions of similar and dissimilar samples are $\left. {C_n^2} \middle/ {C_N^2} \right.$  and  $\left. {(C_N^2-C_n^2)} \middle/ {C_N^2} \right.$ respectively. Total probability of correct and incorrect predictions of certain samples is the proportion of corresponding class, i.e.,
\begin{subequations}\label{eq:2}
\begin{align}
\label{eq:2:1}
\ensuremath{{k_1} + {k_2} &=\left. {C_n^2} \middle/ {C_N^2} \right.,}
\\
\label{eq:2:2}
\ensuremath{{k_3} + {k_4} &= \left. {(C_N^2-C_n^2)} \middle/ {C_N^2} \right..}
\end{align}
\end{subequations}

Based on the given performance of TDGS method about physical similarity, the correctness of data sequences could be analyzed by probability theory. The probability of incorrectly predicting a correct data sequence $\ensuremath{S_0}$ is the union set of incorrectly predicting all similar samples containing $\ensuremath{S_0}$ as dissimilarity, and correctly predicting all dissimilar samples containing $\ensuremath{S_0}$. For the validation set from one discharge, the amounts of similar and dissimilar samples containing $\ensuremath{S_0}$ are $\ensuremath{n-1}$ and $\ensuremath{N-n}$ respectively. The probability of predicting similar samples as dissimilarity is  $1 - {Q_2}$. And the probability of correctly predicting dissimilar samples is  ${Q_3}$.  Considering the proportion of correct data is  ${Q_1}$, the probability of incorrectly predicting correct data $P(R \rightarrow W)$ equals  ${Q_1}{(1 - {Q_2})^{n - 1}}{({Q_3})^{N - n}}$. Since the total probability of correct and incorrect predictions of correct data sequences is  ${Q_1}$, the probability of correctly predicting correct data $P(R \to R)$ equals  ${Q_1} - P(R \to W) = {Q_1}[1{\rm{ - }}{(1 - {Q_2})^{n - 1}}{({Q_3})^{N - n}}]$. The probability of correctly predicting incorrect data sequence $\ensuremath{S_1}$ is the union set of predicting all dissimilar samples containing $\ensuremath{S_1}$ as dissimilarity. In view that the amount of dissimilar samples containing $\ensuremath{S_1}$ is $\ensuremath{N-1}$ and the proportion of incorrect data sequences is   $1 - {Q_1}$, the probability of correctly predicting incorrect data $P(W \to W)$ equals $(1 - {Q_1}){({Q_3})^{N - 1}}$. Since the proportion of incorrect data sequences is $1 - {Q_1}$, the probability of incorrectly predicting incorrect data $P(W \to R)$ equals  $(1 - {Q_1})[1 - {({Q_3})^{N - 1}}]$.

\section{Preference$-$based performance measures for TDGS method in several common data cleaning tasks}
\label{sec:3}

Based on the mapping relations between performance of TDGS method about physical similarity and the correctness of data sequences, performance measures for several common data cleaning tasks are set in this section. 

Different data cleaning tasks have various preferences. Some tasks require high recall rate of incorrect data. Then the performance measure can be set as
\begin{equation}
\label{eq:3}
\begin{split}
\ensuremath{{E_1} = P(W \to W)./[P(W \to W) + P(W \to R)]{\rm{ = }}{({Q_3})^{N - 1}}.}
\end{split}
\end{equation}

Some tasks require high precision of incorrect data. Then the performance measure can be set as
\begin{equation}
\label{eq:4}
\begin{split}
\ensuremath{{E_2} = P(W \to W)./[P(W \to W) + P(R \to W)] = \frac{{1 - {Q_1}}}{{1 - {Q_1} + {Q_1}{{(1{\rm{ - }}{Q_2})}^{n - 1}}{{({Q_3})}^{1 - n}}}}.}
\end{split}
\end{equation}

Some tasks require high recall rate of correct data. Then the performance measure can be set as
\begin{equation}
\label{eq:5}
\begin{split}
\ensuremath{{E_3} = P(R \to R)./[P(R \to R) + P(R \to W){\rm{ = }}1{\rm{ - }}{(1{\rm{ - }}{Q_2})^{n - 1}}{({Q_3})^{N - n}}.}
\end{split}
\end{equation}

Some tasks require high precision of correct data. Then the performance measure can be set as
\begin{equation}
\label{eq:6}
\begin{split}
\ensuremath{{E_4} = P(R \to R)./[P(R \to R) + P(W \to R)] = \frac{{{Q_1}[1{\rm{ - }}{{(1{\rm{ - }}{Q_2})}^{n - 1}}{{({Q_3})}^{N - n}}]}}{{{Q_1}[1{\rm{ - }}{{(1{\rm{ - }}{Q_2})}^{n - 1}}{{({Q_3})}^{N - n}}]{\rm{ + (1 - }}{Q_1}{\rm{)}}[1 - {{({Q_3})}^{^{N - 1}}}]}}.}
\end{split}
\end{equation}

\begin{figure}[htbp]
\centering 
\includegraphics[scale=0.25]{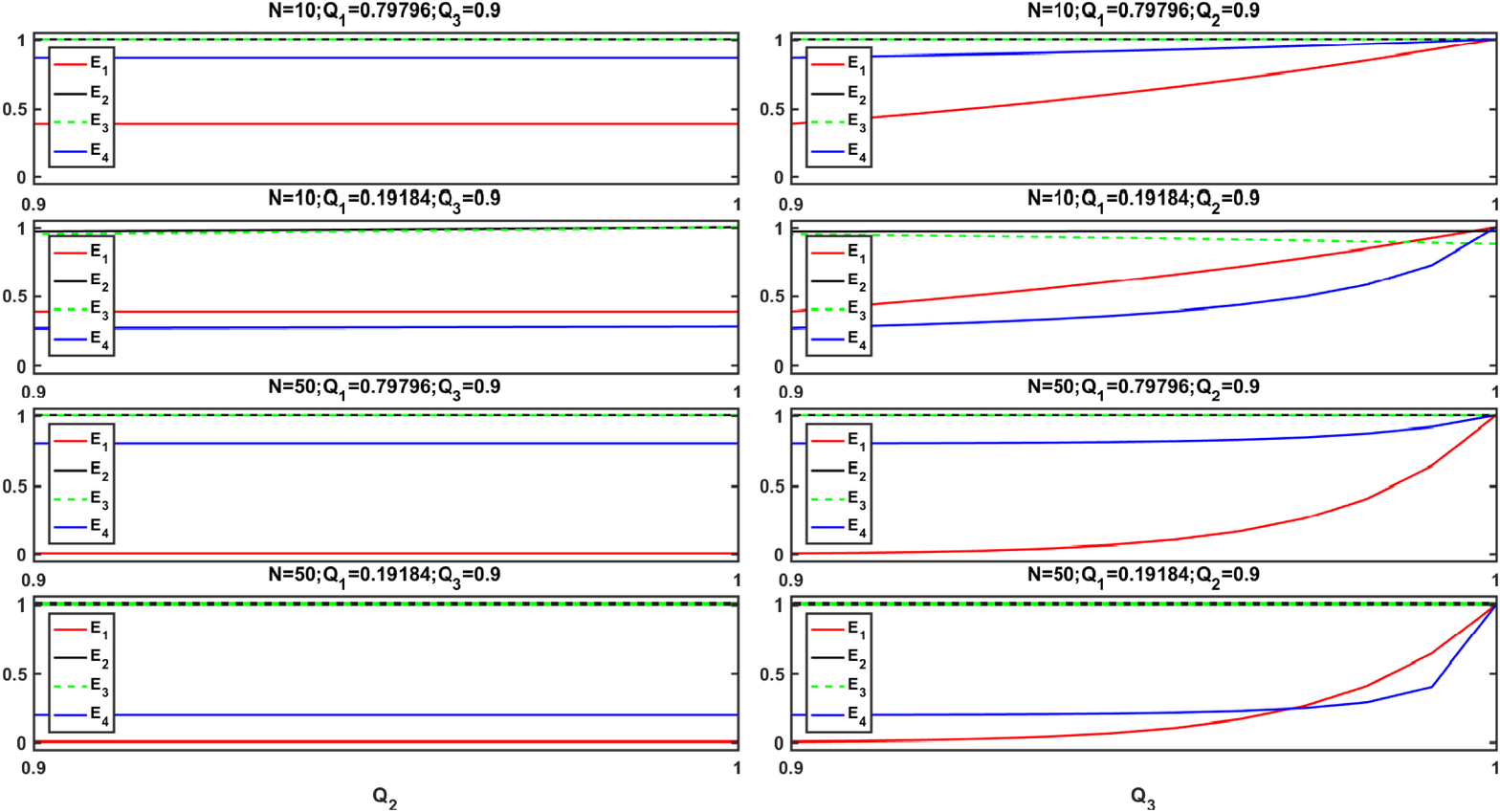}
\caption{\label{fig:1} The change relations between performance parameters about physical similarity and preference-based performance measures are plotted. ${Q_2}$  denotes recall rate of similar samples.  ${Q_3}$ denotes recall rate of dissimilar samples. ${E_1}$  denotes recall of incorrect data.  ${E_2}$ denotes precision of incorrect data.  ${E_3}$ denotes recall of correct data. ${E_4}$  denotes precision of correct data.}
\end{figure}

The change relations between performance parameters about physical similarity and preference-based performance measures are different in various cases. In the case shown in figure~\ref{fig:1}, the recall of incorrect data ${E_1}$ and precision of correct data ${E_4}$ are positive correlative with the recall rate of dissimilar samples ${Q_3}$. In the model selection and evaluation process of this case, the recall of incorrect data and precision of correct data could also be enhanced by just improving the recall rate of dissimilar samples. Then the performance measures ${E_1}$  and ${E_4}$ can be replaced with the more simplified parameter ${Q_3}$. When the channel number of MUM systems is bigger ${(N=50)}$, or the proportion of incorrect data is higher ${(Q_1=0.19)}$, this simplification is more reasonable for the correlative relations between ${Q_3}$  and performance measures are stronger.

\section{Summary}
\label{sec:4}

Data cleaning tasks could be performed better by setting preference-based performance measures. In this paper, we provide the mapping relations between performance of TDGS method about physical similarity and correctness of data sequences by probability theory. Based on these mapping relations, preference-based performance measures for several common data cleaning tasks are set for TDGS method. Meanwhile, the correlative relations between these new performance measures and performance parameters are analyzed.

By setting preference$-$based performance measures, the preferences of data cleaning tasks could be best meet by TDGS method in probability. When these new performance measures are strong positive correlative with certain parameter, preference-based performance measures could be simplified. Next step, we would further improve the performance of TDGS method by adopting different rules for judging the correctness of data based on given physical similarity. The rule adopted in this paper is an absolute judging rule. Next step, we could adopt a non-absolute judging rule. For example, the sequence which is dissimilar from $90\%$ of the other sequences can be tagged as incorrect data. The degree parameter introduced by the judging rule changes the mapping relations between performance of TDGS method about physical similarity and correctness of data sequences. In some cases, proper setting of the degree parameter would improve the data cleaning performance of TDGS method.

\acknowledgments

This research is supported by Key Research Program of Frontier Sciences
CAS (QYZDB-SSW-SYS004), National Natural Science Foundation of China
(NSFC-11575185,11575186), National Magnetic Confinement Fusion Energy
Research Project (2015GB111003,2014GB124005), JSPS-NRF-NSFC A3 Foresight
Program (NSFC-11261140328),and the GeoAlgorithmic Plasma Simulator
(GAPS) Project. 

% We suggest to always provide author, title and journal data:
% in short all the informations that clearly identify a document.

\bibliographystyle{apsrev}
\bibliography{reference1}

\end{document}